\documentclass{article}
\usepackage[dvips]{graphicx}
\usepackage[latin1]{inputenc}
\usepackage{amssymb,amsmath,array}

\usepackage{float}

\let\OLDthebibliography\thebibliography
\renewcommand\thebibliography[1]{
  \OLDthebibliography{#1}
  \setlength{\parskip}{0pt}
  \setlength{\itemsep}{0pt plus 0.3ex}
}

\begin{document}
\title{Feature Relevance Bounds for\\ Ordinal Regression}
\author{Lukas Pfannschmidt$^{1}$, Jonathan Jakob$^{1}$, Michael Biehl$^2$,\\ Peter Tino$^3$, Barbara Hammer$^{1}$
%
\vspace{.3cm}\\
%
1 - Machine learning group,
Bielefeld University, DE\\
2 - Intelligent Systems Group, University of Groningen, NL\\
3 - Computer Science, University of Birmingham, UK
}

\maketitle

Preprint of the ESANN 2019 paper \cite{PfannschmidtFeatureRelevanceBounds2019} as provided by the authors.\\
The original can be found at the ESANN electronics proceedings page.

\begin{abstract} 
The increasing occurrence of ordinal data, mainly sociodemographic, led to a renewed research interest in ordinal regression, i.e.\ the prediction of ordered classes.
Besides model accuracy, 
the interpretation of these models itself is of high relevance, and  existing approaches therefore enforce e.g.\ model sparsity. 
For high dimensional or highly correlated data, however, this might be misleading due to strong variable dependencies. In this contribution, we aim for an identification of feature relevance bounds which -- besides identifying all relevant features --  explicitly differentiates between strongly and weakly relevant features.%
\footnote{Funding by the DFG in the frame of the graduate school DiDy (1906/3) and by the BMBF (grant number 01S18041A) is gratefully acknowledged.}
\end{abstract}

\section{Introduction}
Ordinal data often occur in sociodemographic, financial or medical contexts where it is  hard to give absolute qualitative measurements, but it is easily possible to compare samples.
Another popular ranking on ordinal scales takes place in customer feedback or product ranking by humans \cite{harper2016movielens}.
The {\it ordinal regression problem}~(ORP) is the task to embed given data in the real numbers such that they are ordered according to their label i.e.\ the target variable. 
Although the problem can be attempted with a regular regression or classification method,
dedicated techniques are to be preferred, which can account for the fact that the 
distance between ordinal classes in the data is unknown and not necessarily evenly distributed.
Examples of ordinal regression include
treatments as  multiclass classification problem \cite{FrankSimpleApproachOrdinal2001},
and extensions of standard models such as the support vector machine (SVM) or learning vector quantization (LVQ) to ordinal regression tasks 
\cite{ShashuaRankingLargeMargin,svmo,pOGMLVQ,tino}

Besides a mere classification prescription, practitioners are often interested in the relevance of input features i.e.\ the relevance of ordinal explanatory variables for the given task. This is particularly relevant when the objective exceeds mere diagnostics, such
as safety-critical decision making, or the design of repair strategies. There do exist a few approaches which address 
such feature selection for  ordinal regression:
The approach \cite{GengFeatureSelectionRanking2007} uses a minimal redundancy formulation based on a feature importance score to find the subset of relevant features.
The work in \cite{BaccianellaFeatureSelectionOrdinal2010} focuses on multiple filter methods which are  adapted to ranking data.
These models deliver sparse ordinal regression models which enable some insight into the underlying classification prescription. Yet, their result is arbitrary in the case of correlated and redundant features, where there does not exist a unique minimum relevant feature set. 

The so-called  \emph{all} relevant feature selection problem deals with the challenge to determine all features, which are potentially relevant for a given task -- a problem which is particularly relevant for diagnostics purposes if it is not priorly clear which one of a set of relevant, but redundant features to choose. 
Finding this subset is generally computationally intractable.
For standard classification and regression schemes, a few  efficient
heuristics have recently been proposed:
Some popular models extend predictive models with statistical tests to discriminate between relevance and irrelevance \cite{KursaFeatureSelectionBoruta2010}. 
Another recent approach focuses on the relevant case of linear mappings and phrases the problem to determine the interval of possible variable relevances as a linear optimization problem  \cite{christina}. 

In this paper we introduce an extension of the feature-relevance-interval-computation scheme as proposed in \cite{christina} to  the context of ordinal regression data. For this purpose, we recapture a large margin ordinal regression 
formalization in section \ref{sec:methods}. This is extended to an optimization scheme to determine feature relevance bounds in section \ref{sec:relev_bounds},  which can be transferred to a linear programming problem.
In section \ref{sec:experiments} we  compare our method to classical approaches for artificial data with known ground truth and real life benchmarks.

\section{Large Margin Ordinal Regression}
\label{sec:methods}
Given ordered class labels $L=\{1,2,\ldots,l\}$.
Given training data $X=\{\mathbf x_i^j\in\mathbb{R}^n\:|\: i=1,\ldots,m_ i,\, j\in L\}$ where data point $x_i^j$ is assigned the class label $j\in L$.
The full data set has size $m:=m_1+\ldots+m_l$.
Here the index $j$ refers to the ordinal target variable (represented by $b_j$) the  data point $x_i^j$  belongs to.
The ORP can be phrased as the search for a mapping  $f:\mathbb{R}^n\to\mathbb{R}$ such that 
$f(\mathbf x_{i_1}^{j_1})<f(\mathbf x_{i_2}^{j_2})$ for all $i_1$, $i_2$ and all class labels $j_1<j_2$.

We will restrict to the case of a linear function, i.e.\ $f(\mathbf x) = \mathbf w^t\mathbf x$ with parameter $\mathbf w\in\mathbb{R}^n$.
A popular formulation which is inspired by support vector machines imposes a margin for the embedding  \cite{svmo}:
\begin{eqnarray}
\min_{\mathbf w,b_j,\chi_i^j,\xi_i^j} && 0.5\cdot \|\mathbf w\|_1 + C\cdot \sum_{i,j}\left(\chi_i^j+\xi_i^j\right) \label{eq1}\\
\mbox{s.t.\ for all } i,j&& \mathbf w^t\mathbf x_i^j\le b_j -1+\chi_i^j \label{eq2}\\\notag
&&\mathbf w^t\mathbf x_i^{j+1}\ge b_j+1-\xi_i^{j+1}\\\notag
&&b_j\le b_{j+1}\\\notag
&&\chi_i^j\ge 0, \xi_i^j\ge 0 
\end{eqnarray}
where $\chi_i^j$ and $\xi_i^j$ are slack variables, and the thresholds $b_j$ for $j=1,\ldots, l-1$ determine the boundaries which separate  $l$ classes. The hyper-parameter $C>0$ controls the trade off of the margin and number of errors and can be chosen through cross validation.
Unlike \cite{svmo}, which uses $L_2$ regularisation,
we will use $L_1$ regularisation in (Eq.~\ref{eq1}), aiming for sparse solutions.

%
\section{Feature Relevance Bounds for Ordinal Regression}
\label{sec:relev_bounds}
Assume a training set $X$. Denote an optimum solution of problem (\ref{eq1}) as  $(\tilde{\mathbf w},\tilde b_j,\tilde \xi_i^j, \tilde \chi_i^j)$. This solution induces the
value $\mu_X:=\|\tilde{\mathbf w}\|_1 +C\cdot \sum_{i,j} \left(\tilde \chi_i^j+\tilde \xi_i^j\right)$, which is uniquely determined by $X$.
We are interested  in the {class of equivalent good hypotheses}, i.e.\ all weight vectors $\mathbf{w}$ which yield (almost) the same quality as regards the regression error and generalisation ability as the function induced by $\tilde{\mathbf{w}}$.
For the sake of feasibility, 
we use the following proxy induced by $\mu_X$
\begin{eqnarray}\notag
F_{\delta}(X)&:=&\{\mathbf w\in\mathbb{R}^n\:|\:\exists \xi_i^j, \chi_i^j, b_j \mbox{ such that constraints (\ref{eq2}) hold,}\\\label{eq4}
&& \|\mathbf w\|_1+ C\cdot
\sum_{i,j}\left(\xi_i^j+\chi_i^j\right) \le (1+\delta)\cdot \mu_X\}
\end{eqnarray}
These constraints ensure: 1.~The empirical error of equivalent functions in $F_{\delta}(X)$ is minimum, as measured by the slack variables. 2.~The loss of the generalisation ability  is limited, as guaranteed by a small $L_1$-norm of the 
weight vector and learning theoretical guarantees as provided e.g.\ by Theorem 7 in \cite{agarwal} and Corollary 5 in \cite{zhang}.
The parameter $\delta\ge 0$ quantifies the tolerated deviation to accept a function as yet good enough, C is chosen according to the solution of Problem (\ref{eq1}).

Solutions $\mathbf{w}$ in $F_{\delta}(X)$ are sparse in the sense that irrelevant features are uniformly weighted as $0$ for all solutions in $F_{\delta}(X)$. Relevant but potentially redundant features can be weighted arbitrarily, disregarding sparsity, similar in spirit to 
the elastic net, which weights mutually redundant  features equally \cite{elasticnet}.
In this contribution we are interested in the 
relevance of features for forming good hypotheses, where we 
are interested in the following more specific characteristics:
\begin{itemize}
\item {\bf Strong relevance} of feature $I$ for $F_{\delta}(X)$: Is feature $I$ relevant for all hypotheses in $F_{\delta}(X)$, i.e.\ all
weight vectors $\mathbf w\in F_{\delta}(X)$ yield $w_I\not= 0$?
\item {\bf Weak relevance} of feature $I$ for $F_{\delta}(X)$: Is feature $I$ relevant for at least one hypothesis in $F_{\delta}(X)$ in the sense that one 
weight vector $\mathbf w\in F_{\delta}(X)$ exists with $w_I\not= 0$, but this does not hold for all weight vectors in $F_{\delta}(X)$?
\item {\bf Irrelevance} of feature $I$ for $F_{\delta}(X)$: Is feature $I$ irrelevant for every hypothesis in $F_{\delta}(X)$, i.e.\ all
weight vectors $\mathbf w\in F_{\delta}(X)$ yield $w_I= 0$?
\end{itemize}
A feature is irrelevant for $F_{\delta}(X)$ if it is neither strongly nor weakly relevant.
The questions of strong and weak relevance can be answered via the following optimisation problems:
\begin{description}
\item[Problem $\mathbf{minrel}(I)$:]
\begin{eqnarray}
\min_{\mathbf w,b_j,\chi_i^j,\xi_i^j} && |w_I| \label{eq5}\\\notag
\mbox{s.t.\ for all } i,j&& \mbox{conditions (\ref{eq2}) hold and}\\
&&\|\mathbf{w}\|_1 + C\cdot \sum_{k,l}\left(\chi_k^l+\xi_k^l \right)\le (1+\delta)\cdot\mu_X \label{eq6}
\end{eqnarray}
Feature $I$ is strongly relevant for $F_{\delta}(X)$ iff $\mathrm{minrel}(I)$ yields an optimum larger than $0$.
\item[Problem $\mathbf{maxrel}(I)$:]
\begin{eqnarray}
\max_{\mathbf w,b_j,\chi_i^j,\xi_i^j} && |w_I| \label{eq8}\\\notag
\mbox{s.t.\ for all } i,j&& \mbox{conditions (\ref{eq2}) hold and}\\\notag
&&\|\mathbf{w}\|_1 + C\cdot \sum_{k,l}\left(\chi_k^l+\xi_k^l \right)\le (1+\delta)\cdot\mu_X 
\end{eqnarray}
Feature $I$ is weakly relevant for $F_{\delta}(X)$ iff $\mathrm{minrel}(I)$ yields an optimum $0$
and $\mathrm{maxrel}(I)$ yields an optimum larger than $0$
\end{description}
These two optimisation problems span a real-valued interval for every feature $I$ with the result of 
$\mathrm{minrel}(I)$ as lower and $\mathrm{maxrel}(I)$ as upper bound. This interval characterises the range 
of weights for $I$ occupied by good solutions in $F_{\delta}(X)$. Hence, besides information about a features relevance,
some indication about the degree up to which a feature is relevant or can be substituted by others, is given.
Note, however, that the solutions are in general not consistent estimators of an underlying \lq true\rq\ weight vector as regards its exact value, 
as has been discussed e.g.\ for Lasso \cite{consistencylasso}. 
Similar to \cite{christina}, these optimization problems can be rephrased as equivalent linear optimization problems which can be solved efficiently in polynomial time. Here, we omit this formulation and proof due to page limitations.

\paragraph*{{\bf Threshold Selection:}} 
\label{sub:handling_numerical_instabilities}
To estimate a threshold for a feature to be considered as weakly relevant, we generate features with the same statistical properties but no relevance by design, sometimes referred to as probe or shadow features \cite{StoppigliaRankingrandomfeature2003}:
We permute feature $I$ and compute the relevance interval bounds using  $I_P$ instead of $I$. We repeat this $d$ times. For a cutoff, we
accept 
a certain rate  $r_{FP}:=0.01$ of false positives.
From the descending list of all upper probe interval bounds we pick the threshold  with index $\lfloor r_{FP}\times d\rfloor$.
            
To determine, if a feature is strongly relevant, it is sufficient to check whether the problem for the lower bound and feature $I_P$ is  deemed infeasible by the solver,
since the accuracy of the model without  feature $I$ degrades significantly.

\section{Experiments}
\label{sec:experiments}

\paragraph*{Artificial Data}
  \label{subs:artif_data}
  We adapt the generation method presented in \cite{christina} for ordinal regression:
We create random values to define a  hyperplane  - the variables correspond to strongly relevant features.
  Weakly relevant features are created by replacing some of the strongly relevant values with a correlated linear combination. 
  The hyperplane can be used to sample continuous values.
  By using equal frequency binning we converted the continuous regression variable into an ordered discrete target variable.
  Gaussian noise as well as additional noisy features are added.
 Several data sets with different characteristics as regards the amount of strongly, weakly and irrelevant variables
 are created this way, see.
 Table~\ref{t:toy_set_params}.

 For evaluation,  we consider cross-validated feature elimination.
We compare our model (dubbed feature relevance interval - FRI)\footnote{ Implementation in Python: https://github.com/lpfann/fri} to  ordinal regression with  the Lasso penalty \cite{tibshirani_regression_1996} and  the ElasticNet \cite{ZouRegularizationvariableselection2005} penalty with ratio of $0.5L_1 + 0.5L_2$. Hyperparameters are selected according to 5-fold cross validation.
We evaluate whether the method is able to identify all relevant features, whereby we evaluate the correspondence of the sets by the true and detected set of relevant features by the F-measure, see Table~\ref{tbl:toyperf}.
In all cases where weakly relevant features are involved, FRI provides significantly better accuracy than elastic net and Lasso to detect all relevant features. 

  \begin{table}
  \caption{Artificially created data sets with known ground truth and evaluation of the identified  relevant features by the methods as compared to all relevant features. The score is averaged over 30 independent runs.}
  
  ~\\
  \label{t:toy_set_params}\label{tbl:toyperf}
  \centering
  \begin{tabular}{c c c c || c c c} 
  \hline
  \hline
  \multicolumn{4}{c||}{Data Set Characteristics}&\multicolumn{3}{c}{Results (F-measure)}\\ \hline
  
{\textit{Points}} & {\textit{Strong}} & {\textit{Weak}} & {\textit{Irrelevant}} &
  \emph{ElasticNet}  & \emph{FRI}   & \emph{Lasso} \\
  \hline
 150 & 6 & 0 & 6  & \textbf{0.91}        & 0.87           & 0.88      \\
 150 & 0 & 6 & 6  & 0.67                 & \textbf{0.93}  & 0.64  \\
 150 & 3 & 4 & 3  & 0.85                 & \textbf{0.91}  & 0.83     \\
256 & 6 & 6 & 6   & 0.88                 & \textbf{0.97}  & 0.87   \\
 512 & 1 & 2 & 11  & 0.81                 & \textbf{0.87}  & 0.76    \\
  200 & 1 & 20 & 0  & 0.36                 & \textbf{1.00}  & 0.26    \\
 200 & 1 & 20 & 20  & 0.36                 & \textbf{0.90}  & 0.41     \\

  \hline
  \hline

  \end{tabular}
  \end{table}

\paragraph*{Benchmark Data}
We  test the proposed method on  benchmark data as described in   \cite{realOrdinalData}. These data sets are imbalanced.
  The proposed model provides a mapping $\mathbf{x}\to f(\mathbf x) = y\in\{1,\ldots,l\}$
  where $f(\mathbf{x})=\mathrm{argmin}_i \{\mathbf{w}^t\mathbf{x}\le b_i\}$ where $b_l:=\infty$. 
The output is evaluated by the
   macro-averaged absolute error: $\mathrm{MMAE} =  \sum^l_j\frac{\sum_{i} |j-f(\mathbf{x}_i^j)|}{m_j}/l$ where $m_j$ is the number of examples in class $j$.
  We replicated the experiments which have been presented in   \cite{pOGMLVQ,tino}, whereby results are averaged over 30 folds. 
  Results are reported in Tab.~\ref{t:bench_result}.
  Since all methods rely on linear models, their MMAE is comparable. For these data, no ground truth as regards the relevant features is available, so we can only compare the amount of information provided by the methods. We report the average number of features identified as relevant by the techniques. For three data sets (Squash-stored, Squash-unstored, TAE), FRI identifies a smaller number of relevant features than the alternatives, yielding the same accuracy. For three further data sets (Automobile, Eucalyptus, Pasture), FRI identifies more (weakly relevant) features. In all cases, FRI  offers more information than direct Lasso or ElasticNet by identifying weakly relevant features.




    \begin{table}
      \caption[Experimental Results]{Left: MMAE of FRI, elastic net (EN) and Lasso  along with standard deviations ($\pm$) across 30 folds of data sets from \cite{ZouRegularizationvariableselection2005,realOrdinalData}. Right: Mean feature set size of the methods. FRI allows extra discrimination between strong ($FRI_s$) relevant and weak ($FRI_w$) relevance.\\}
      \label{t:bench_result}
      \centering
        \begin{tabular}{l|c||rr|rr}
        \hline\hline
        \multicolumn{1}{l|}{}      & \multicolumn{1}{c||}{} & \multicolumn{4}{c}{Average Feature Set Size}  \\ 
                                  &      MMAE            & $FRI_s$  & $FRI_w$  & EN   & Lasso         \\ 
        \hline
        \textbf{Automobile }      & 0.661 $\pm$ 0.129    & 4.5      & 12.6     & 4.0  & 4.8           \\
        \textbf{Bondrate }        & 1.36 $\pm$ 0.122     & 0.0      & 5.4      & 2.0  & 2.0           \\
        \textbf{Contact-lenses }  & 0.914 $\pm$ 0.206    & 0.9      & 1.1      & 2.0  & 2.0           \\
        \textbf{Eucalyptus }      & 0.406 $\pm$ 0.027    & 2.1      & 33.2     & 15.6 & 15.7          \\
        \textbf{Newthyroid }      & 0.667 $\pm$ 0.0      & 0.0      & 4.7      & 2.0  & 2.0           \\
        \textbf{Pasture }         & 0.367 $\pm$ 0.121    & 0.0      & 15.5     & 6.0  & 5.1           \\
        \textbf{Squash-stored }   & 0.39 $\pm$ 0.164     & 2.4      & 7.9      & 11.1 & 6.6           \\
        \textbf{Squash-unstored}  & 0.317 $\pm$ 0.168    & 1.8      & 3.3      & 8.0  & 7.3           \\
        \textbf{TAE }             & 0.621 $\pm$ 0.153    & 1.9      & 5.4      & 16.8 & 13.7          \\
        \textbf{Winequality-red}  & 1.081 $\pm$ 0.037    & 0.0      & 7.6      & 5.4  & 5.3           \\
        \hline\hline
        \end{tabular}

  \end{table}

\section{Conclusions}
  \label{sec:conclusion}

  In this paper we presented the adaption of the feature relevance bounds approach to ordinal regression data.
  Based on the experiments we showed that the method can provide a good all-relevant feature set approximation in this new setting.
  These feature sets represent additional information useful in analytic use cases for model and experiment design, subject for further evaluation.

{
\footnotesize

\bibliographystyle{unsrt}
\bibliography{ordinal.bib}

}

\end{document}